\documentclass[sigconf]{acmart}
\AtBeginDocument{%
  }

\copyrightyear{2026}
\acmYear{2026}
\setcopyright{cc}
\setcctype{by}

\acmConference[MM '26]
{Proceedings of the 34th ACM International Conference on Multimedia}
{November 10--14, 2026}
{Rio de Janeiro, Brazil.}

\acmBooktitle{Proceedings of the 34th ACM International Conference on Multimedia (MM '26), November 10--14, 2026, Rio de Janeiro, Brazil}
\acmISBN{979-8-4007-2213-4/2026/11}
\acmDOI{10.1145/3767308.3835040}

\usepackage{multirow}
\usepackage{booktabs}
\usepackage{xcolor}
\usepackage{microtype}
\usepackage{balance}
\newcommand{\framecredit}[1]{%
  \textit{Source frames and subtitle excerpts: #1\unskip.
  Copyright remains with the respective original video owners;
  reproduced in low resolution under fair use for scholarly analysis.}%
}

\begin{document}

\title[Where to Focus: Query-Modulated Keyframe Selection]{Where to Focus: Query-Modulated Multimodal Keyframe Selection for Long Video Understanding}

\author{Shaoguang Wang}
\orcid{0009-0000-4956-784X}
\affiliation{%
  \institution{The Hong Kong University of Science and Technology (Guangzhou)}
  \city{Guangzhou}
  \country{China}
}
\email{swang440@connect.hkust-gz.edu.cn}

\author{Weiyu Guo}
\orcid{0000-0001-5449-9490}
\affiliation{%
  \institution{The Hong Kong University of Science and Technology (Guangzhou)}
  \city{Guangzhou}
  \country{China}
}
\email{guoweiyu96@gmail.com}

\author{Ziyang Chen}
\orcid{0009-0006-8895-1805}
\affiliation{%
  \institution{The Hong Kong University of Science and Technology (Guangzhou)}
  \city{Guangzhou}
  \country{China}
}
\email{zchen483@connect.hkust-gz.edu.cn}

\author{Xuming Hu}
\correspondingauthor
\orcid{0000-0001-6075-4224}
\affiliation{%
  \institution{The Hong Kong University of Science and Technology (Guangzhou)}
  \city{Guangzhou}
  \country{China}
}
\affiliation{%
  \institution{The Hong Kong University of Science and Technology}
  \city{Hong Kong SAR}
  \country{China}
}
\email{xuminghu@hkust-gz.edu.cn}

\author{Hui Xiong}
\correspondingauthor
\orcid{0000-0001-6016-6465}
\affiliation{%
  \institution{The Hong Kong University of Science and Technology (Guangzhou)}
  \city{Guangzhou}
  \country{China}
}
\affiliation{%
  \institution{The Hong Kong University of Science and Technology}
  \city{Hong Kong SAR}
  \country{China}
}
\email{xionghui@ust.hk}

\renewcommand{\shortauthors}{Shaoguang Wang, Weiyu Guo, Ziyang Chen, Xuming Hu, and Hui Xiong}

\begin{abstract}
Long video understanding remains a formidable challenge for Multimodal Large Language Models (MLLMs) due to the prohibitive cost of processing dense frame sequences. Prevailing keyframe-selection methods rely on either a single visual-centric metric (e.g., CLIP similarity) or a static fusion of heuristic scores. This ``one-size-fits-all'' paradigm frequently fails: visual-only metrics are ineffective for plot-driven narrative queries, while indiscriminately adding textual scores introduces severe ``modal noise'' for purely visual tasks. To break this bottleneck, we propose \textbf{Q-Gate}, a plug-and-play, training-free framework that treats keyframe selection as a dynamic modality routing problem. We decouple retrieval into three lightweight expert streams: \textit{Visual Grounding} for local details, \textit{Global Matching} for scene semantics, and \textit{Contextual Alignment} for subtitle-driven narratives. Crucially, Q-Gate introduces a \textbf{Query-Modulated Gating Mechanism} that uses the in-context reasoning of an LLM to assess query intent and dynamically allocate weights across the experts, activating necessary modalities while ``muting'' irrelevant ones to maximize the signal-to-noise ratio. Extensive experiments on LongVideoBench and Video-MME across
multiple MLLM backbones show that Q-Gate outperforms
representative keyframe-selection baselines in most settings,
with particularly strong gains on long and medium videos,
providing a robust and interpretable solution for scalable
video reasoning.
\end{abstract}



\begin{teaserfigure}
  \centering
  \includegraphics[width=\textwidth]{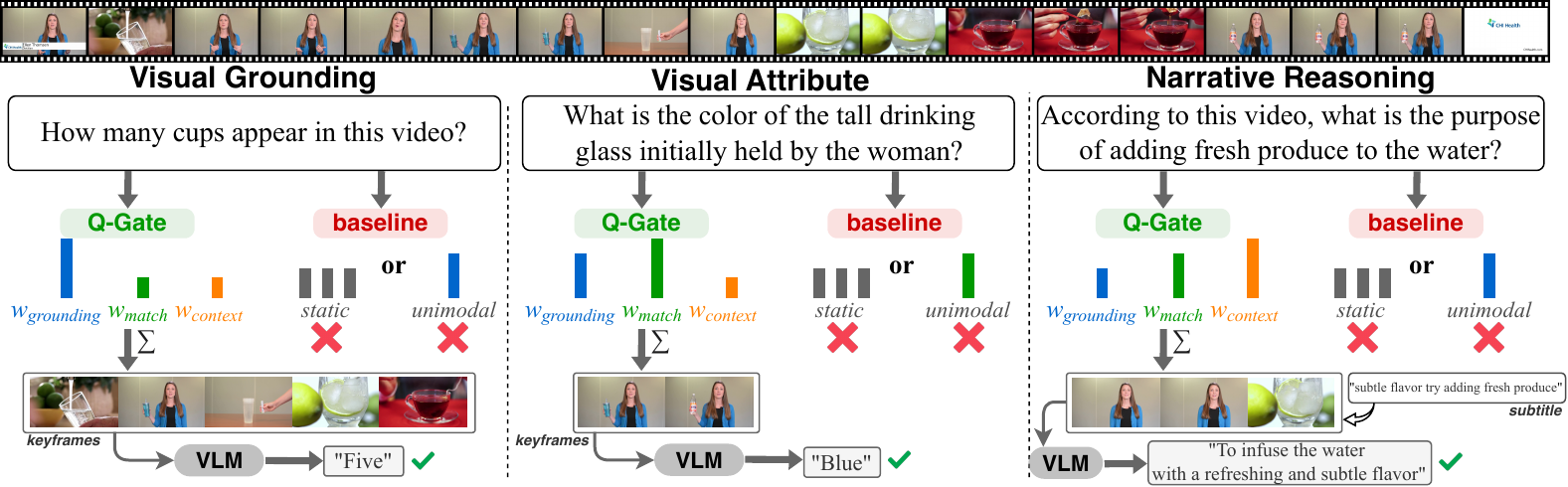} 
  \caption{\textbf{The Q-Gate in Action: Same Video, Different Intents.}
  Static or unimodal strategies (gray bars) fail on complex multimodal queries. Q-Gate instead adapts to the query's intent (colored bars), prioritizing different expert streams to know when to ``look'' and when to ``listen.''
  \framecredit{
    Video-MME~\cite{Fu_2025_CVPR},
    video ID \texttt{540LkURTR7g}
  }
  }
  \Description{For one video and three different questions, a comparison showing that a fixed strategy selects the wrong keyframes while Q-Gate reweights its expert streams per question and selects the frames that yield the correct answer.}
  \label{fig:teaser}
\end{teaserfigure}

\maketitle

\section{Introduction}
\label{sec:intro}

The advent of Multimodal Large Language Models (MLLMs) has revolutionized video understanding, enabling complex reasoning over visual and textual data~\cite{10982110}. However, the leap from short clips to long videos, spanning minutes to hours, unveils a critical bottleneck: the prohibitive computational cost and token limitations of MLLMs~\cite{10.1007/978-3-031-73414-4_26, park-etal-2026-many}. Processing thousands of frames is infeasible, forcing models to rely on a sparsely sampled subset of keyframes. Consequently, the quality of this subset fundamentally dictates the upper bound of the model's comprehension capabilities. This raises a pivotal question: \textit{How can we efficiently select a minimal yet comprehensive set of keyframes that are most relevant to the visual and textual cues embedded within a user's query?}

Current keyframe selection strategies, while pioneering, often operate with a unimodal bias or a static fusion logic, failing to capture the dynamic nature of human cognition. On one hand, vision-centric methods excel at identifying visually salient moments. For instance, relevance-based approaches like AKS~\cite{Tang_2025_CVPR} leverage global image-text matching to find semantically similar scenes, while logic-based methods like VSLS~\cite{NEURIPS2025_b432bc78} perform fine-grained entity detection between different frames to verify specific semantic logic relations. Recent exploration-exploitation frameworks like FOCUS~\cite{zhu2026focus} introduce bandit-based uncertainty for sampling. However, these methods act like a ``deaf observer'': they are prone to failure when the query's answer is embedded in the narrative context, such as dialogue or causal explanations found in subtitles (Figure~\ref{fig:framework}, down branch), as they rely on visual-only relevance and cannot adaptively switch between visual and textual information.

On the other hand, recent attempts to incorporate textual information often employ a static integration strategy. This ``look and listen simultaneously'' approach introduces significant \textbf{modal noise}, a form of cross-modal negative transfer, especially when the query is purely visual. For example, as illustrated in Figure~\ref{fig:teaser} (left), when asked to count objects, irrelevant subtitles can distract the model and degrade performance. This reveals a fundamental flaw: existing methods lack a mechanism to dynamically decide \textit{when} to look and \textit{when} to listen. Human cognition, in contrast, is highly adaptive. When asked a visual question, we instinctively focus our visual attention; when asked for a reason, as in Figure~\ref{fig:teaser} (right), we recall the conversation.

To bridge this gap, we propose \textbf{Q-Gate}, a novel, training-free framework that introduces a \textbf{Query-Modulated Gating Mechanism} for keyframe selection. Instead of statically fusing information, Q-Gate first analyzes the user's query to understand its underlying intent. It then adaptively modulates the weights between three complementary perceptual streams: 1) \textbf{Visual Grounding}, which performs object-level verification for fine-grained details; 2) \textbf{Global Matching}, which captures frame-level semantic context; and 3) \textbf{Contextual Alignment}, which localizes narrative cues from subtitles. This ``Look and Listen'' strategy allows Q-Gate to prioritize visual streams for descriptive questions while dynamically shifting focus to the narrative stream for plot-related or reasoning-based queries, thereby maximizing signal-to-noise ratio.

Our contributions are threefold:
\begin{itemize}
    \item We formulate keyframe selection as a query-modulated routing problem, operationalized as a training-free Zero-Shot Mixture-of-Experts system in which an LLM acts as an intelligent gating network to allocate attention across three complementary expert streams.
    \item We design a unified normalization pipeline with a Masked Temperature Softmax and show in a diagnostic study that a rule-based heuristic does not outperform naive static fusion, highlighting the value of semantic reasoning for routing.
    \item Extensive experiments show that Q-Gate outperforms
    representative training-free baselines in most settings, with analyses
    confirming its interpretability, efficiency, and robustness across backbones and frame budgets.
\end{itemize}

\section{Related Work}
\label{sec:related_work}

Our work is positioned at the confluence of three rapidly evolving research areas: the application of Large Language Models (LLMs) to video understanding, advanced strategies for keyframe selection in long-form videos, and the development of dynamic multimodal fusion techniques to enhance cross-modal alignment.

\subsection{Large Language Models for Video Understanding}

The remarkable success of Large Language Models (LLMs)~\cite{devlin-etal-2019-bert, NEURIPS2020_1457c0d6, touvron2023llamaopenefficientfoundation} has catalyzed a paradigm shift towards multimodal intelligence. Vision-Language Models (VLMs), such as LLaVA~\cite{NEURIPS2023_6dcf277e}, InternVL~\cite{Chen_2024_CVPR}, and the Qwen-VL series~\cite{bai2023qwenvlversatilevisionlanguagemodel, bai2025qwen3vltechnicalreport}, have achieved unprecedented performance on image-based tasks by aligning rich visual representations with the semantic space of LLMs.

Extending these capabilities to the temporal dimension of video, however, presents substantial challenges. A primary line of work adapts image-centric architectures for video, creating models like Video-LLaMA~\cite{zhang-etal-2023-video}, Video-LLaVA~\cite{lin-etal-2024-video}, and LLaVA-Interleave~\cite{ICLR2025_c9f95e9e}. While effective for short clips, these models grapple with the quadratic complexity of self-attention when faced with long-form videos, which can contain tens of thousands of frames. To mitigate this ``context-length crisis'', a variety of compression and efficiency-focused strategies have emerged. These include employing memory mechanisms~\cite{Song_2024_CVPR, 10.1007/978-3-031-72670-5_5}, designing efficient transformer architectures like Ring Attention~\cite{ICLR2024_11195878} and Longformer~\cite{beltagy2020longformerlongdocumenttransformer}, and developing token reduction methods~\cite{song-etal-2025-less, NEURIPS2024_871ed095, Wang_2026_CVPR}. Our work is complementary: instead of compressing the visual signal, we select the most salient raw frames, preserving full visual fidelity at critical moments while reducing the token load.

\subsection{Keyframe Selection for Long Videos}

Keyframe selection is a fundamental prerequisite for efficient long video understanding. The goal is to distill a long, redundant video into a concise yet comprehensive set of frames. This problem can be broadly categorized into query-agnostic and query-driven approaches, which differ in their objectives and applicability. Unlike learning-based approaches such as Frame-Voyager~\cite{yu2025framevoyager}, our work concentrates on the training-free methods.

\paragraph{Query-Agnostic Summarization.} Traditional methods often focus on creating a generic summary of the video, independent of any specific query. These techniques range from classic visual feature analysis, such as detecting shot boundaries~\cite{9594911}, to more advanced deep learning approaches that identify visually diverse or representative frames~\cite{NIPS2014_5d3b9e06}. While useful for general-purpose browsing, these summaries are suboptimal for task-specific QA, as they may discard frames that are visually mundane but crucial for answering a specific plot-related question.

\paragraph{Query-Driven Temporal Grounding.} More recent and relevant to our work are query-driven methods, which aim to localize frames pertinent to a user's textual query. This paradigm has evolved along several distinct lines:
\begin{itemize}
    \item \textbf{Relevance-Based Retrieval:} A dominant approach treats frame selection as a retrieval task, where a score is computed for each frame based on its cross-modal similarity to the query, typically using a pre-trained VLM like CLIP~\cite{pmlr-v139-radford21a}. The highest-scoring frames are then selected~\cite{park-etal-2026-many, 11397184}. While powerful for capturing global semantics, this method can overlook fine-grained details. To enhance coverage, some works like AKS~\cite{Tang_2025_CVPR} introduce adaptive partitioning to ensure frames are sampled from different temporal segments, but the core scoring remains vision-centric.
    
    \item \textbf{Logic-Based Verification:} To improve precision, another research direction focuses on decomposing the query into a set of detectable visual entities and logical relationships~\cite{NEURIPS2025_b432bc78}. These frameworks often employ an iterative search process, using open-vocabulary object detectors like YOLO-World~\cite{Cheng_2024_CVPR} to verify the presence of specific objects and relations within frames, thereby refining a sampling distribution. This ``detect-then-verify'' paradigm excels at detail-oriented questions but struggles with abstract concepts or when the visual evidence is ambiguous. T*~\cite{Ye_2025_CVPR} further extends this by reframing temporal search as an iterative spatial search problem.
    
    \item \textbf{Agentic and Hierarchical Search:} The most recent trend involves using an LLM as a reasoning agent to guide the search process. Frameworks like VideoAgent~\cite{10.1007/978-3-031-72989-8_4, 10.1007/978-3-031-72670-5_5} and VideoTree~\cite{Wang_2025_CVPR} employ multi-step, iterative reasoning to decide which frames to analyze next. While demonstrating impressive reasoning capabilities, these iterative approaches often incur significant latency due to multiple sequential calls to the LLM. Q-Gate, in contrast, performs its query analysis and weight modulation in a single, non-iterative pass, achieving a more favorable balance between performance and efficiency.
\end{itemize}

\subsection{Multimodal Fusion in Video QA}

The synergy of multiple modalities is central to comprehensive video understanding. Early fusion techniques often involved concatenating raw features, while late fusion averaged unimodal predictions. Modern approaches leverage more sophisticated cross-modal attention mechanisms to model the intricate interactions between visual, textual, and sometimes audio streams~\cite{li-etal-2020-hero, zhang-etal-2023-video}. Recent advances in multimodal video QA have shown that even a simple, additive combination of subtitle similarity with visual relevance scores can yield measurable improvements, pointing to the untapped potential of narrative context. However, such static
fusion remains query-agnostic and fails to account for the varying
importance of different modalities for different questions. For
instance, a purely visual question about object colors can be
contaminated by irrelevant subtitle scores. In stark contrast, our
Q-Gate framework introduces a dynamic gating mechanism. This can
be viewed as a form of query-conditioned attention over modalities,
where the gater decides when to prioritize visual evidence and when
to shift focus to the narrative stream.

\begin{figure*}
  \centering
  \includegraphics[width=\textwidth]{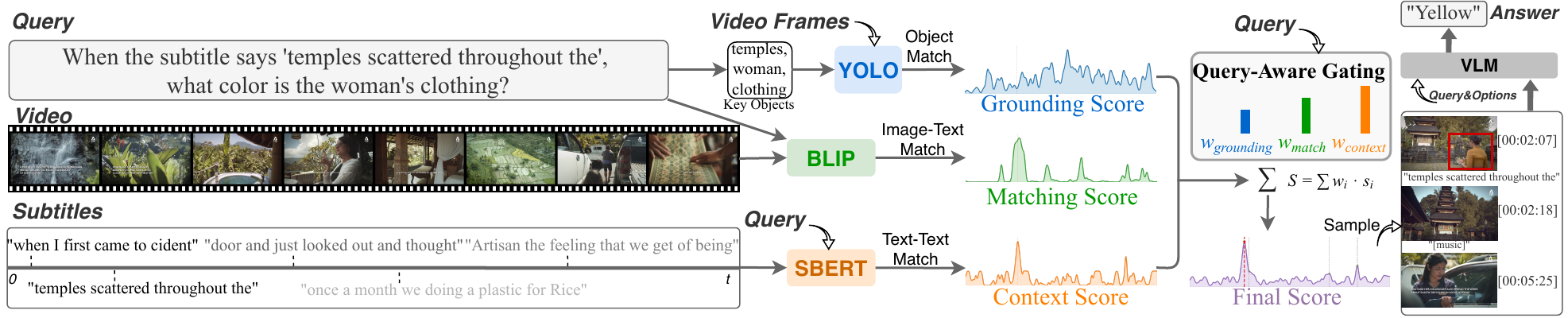}
  \caption{Overview of the \textbf{Q-Gate} framework. Three parallel expert streams compute multi-granularity scores: (a) \textbf{Visual Grounding} for object-level details, (b) \textbf{Global Matching} for scene semantics, and (c) \textbf{Contextual Alignment} for narrative cues. A \textbf{Query-Aware Gating} module modulates these streams into a final distribution, from which the Sampler selects the top-$K$ frames and subtitles for the downstream VLM.
  \framecredit{
  LongVideoBench~\cite{NEURIPS2024_329ad516},
  video ID \texttt{9PD3ciudpIE}
  }
  }
  \Description{A pipeline diagram: a video and query feed three scoring streams (grounding via YOLO, matching via BLIP, contextual via SBERT); an LLM gater produces per-stream weights that are fused into a final score, and a sampler selects the top-K frames and subtitles as the VLM prompt.}
  \label{fig:framework}
\end{figure*}
\section{Methodology}
\label{sec:method}

To address the challenge of selecting query-relevant keyframes, we introduce \textbf{Q-Gate}, a training-free framework that emulates the human cognitive process of adaptively focusing on different information modalities. As illustrated in Figure~\ref{fig:framework}, Q-Gate operates in three main stages: (1) Multi-Granularity Scoring, where parallel streams compute relevance scores from different perspectives; (2) Query-Aware Gating, where a dynamic weighting mechanism determines the importance of each stream based on the user's query; and (3) Sampling \& Inference, where the final keyframes are selected and fed to a downstream VLM.

\subsection{Multi-Granularity Scoring Streams}
\label{ssec:scoring}

Given a long video $V$ with $T$ frames and a textual query $q$, our first step is to generate three complementary, time-aligned score distributions: $S_g, S_m, S_c \in \mathbb{R}^T$. Each stream captures a different facet of relevance, spanning from fine-grained objects to global semantics and narrative context. A unified normalization pipeline is then applied to ensure all three scores are mathematically comparable. 

\subsubsection{Visual Grounding Stream ($S_g$)}
This stream aims to identify frames containing specific, concrete visual entities mentioned in the query. It reflects the ``detective's'' perspective, focusing on ``who,'' ``what,'' and ``where.'' Visual Grounding excels at verifying explicitly mentioned entities but may struggle with abstract concepts, whereas Global Matching provides a holistic semantic fallback. The grounding process involves:
\begin{enumerate}
    \item \textbf{Entity Extraction:} We first parse the query $q$ to extract a set of key visual entities $E_q = \{e_1, e_2, ...\}$ using an LLM, such as ``woman'' or ``red car.''
    \item \textbf{Frame-wise Verification:} For each frame $v_t$, we employ an open-vocabulary object detector YOLO-World~\cite{Cheng_2024_CVPR} to compute a raw grounding score, $s_g^{raw}(t)$. This score represents the maximum confidence of detecting any entity $e_i \in E_q$ within the frame, and also considers the satisfaction of spatial relationships between entities. This results in a raw score vector that highlights moments of high object-level relevance.
\end{enumerate}

\subsubsection{Global Matching Stream ($S_m$)}
While grounding excels at finding specific objects, it may miss the overall scene context. This stream captures the global semantic similarity between each frame and the query, akin to an ``artist's'' perspective on atmosphere.
\begin{enumerate}
    \item \textbf{Joint Encoding:} We use the image-text matching model of BLIP~\cite{pmlr-v162-li22n} (the \texttt{blip\_image\_text\_matching} \texttt{large} variant), which feeds a frame $v_t$ and the query $q$ jointly through a multimodal encoder with an image-text matching (ITM) head.
    \item \textbf{Matching Scoring:} The raw matching score $s_m^{raw}(t)$ is the ITM head's matching probability, i.e., the softmax over its two-way logits taken at the matched class:
    \begin{equation}
        s_m^{raw}(t) = \mathrm{softmax}\big(\mathrm{ITM}(v_t, q)\big)_{\text{match}}.
        \label{eq:matching_score}
    \end{equation}
\end{enumerate}
This stream is robust for general scene understanding but may overlook fine-grained details.

\subsubsection{Contextual Alignment Stream ($S_c$)}
Many questions in long videos, particularly those involving reasoning or plot, cannot be answered by visual information alone. This stream leverages the narrative content from subtitles, reflecting a ``stenographer's'' perspective that focuses on non-visual, dialogue-driven cues.
\begin{enumerate}
    \item \textbf{Temporal Mapping:} For each frame at time $t$, we identify the corresponding subtitle text $sub_t$.
    \item \textbf{Textual Similarity:} We use a sentence-embedding model (e.g., Sentence-BERT~\cite{reimers-gurevych-2019-sentence}) to compute the raw context score $s_c^{raw}(t)$ as the cosine similarity between the query $q$ and $sub_t$. If no subtitle exists, $s_c^{raw}(t) = 0$.
\end{enumerate}
This stream is crucial for plot-related questions but provides no information for purely visual queries.

\subsubsection{Unified Normalization Pipeline}
The raw scores from the three streams ($S_g^{raw}, S_m^{raw}, S_c^{raw}$) have disparate scales and distributions. To make them mathematically comparable for fusion, we apply a unified two-step normalization pipeline to each raw score vector $S_i^{raw}$ independently:
\begin{enumerate}
    \item \textbf{Min-Max Scaling:} We first scale the scores to a common range of $[0, 1]$ to eliminate scale discrepancies:
    \begin{equation}
        S_i^{scaled}(t) = \frac{s_i^{raw}(t) - \min(S_i^{raw})}{\max(S_i^{raw}) - \min(S_i^{raw})}.
    \end{equation}
    When $\max(S_i^{raw}) = \min(S_i^{raw})$ (a degenerate stream with no discriminative signal, e.g., a subtitle-free video), we set $S_i^{scaled}(t) = 0$ for all $t$.
    \item \textbf{Masked Temperature Softmax:} To amplify high-relevance peaks and suppress low-level noise, we apply a softmax with temperature $\tau$. After Min-Max scaling, the positions attaining each stream's minimum are mapped to zero (for Contextual Alignment these are the subtitle-free frames). A standard softmax would still assign them positive probability since $\exp(0)=1$, diluting the mass of genuinely relevant frames. We therefore exclude these zero-valued scaled positions from the normalization so they retain zero probability. The final normalized score is defined as:
    \begin{equation}
        S_i(t) = \begin{cases}
                    \frac{\exp(S_i^{scaled}(t) / \tau)}{\sum_{j:S_i^{scaled}(j)>0} \exp(S_i^{scaled}(j) / \tau)} & \text{if } S_i^{scaled}(t) > 0 \\
                    0 & \text{otherwise}
                \end{cases}
    \end{equation}
    where we use $\tau=0.5$ in our experiments to create a sharpened distribution that is favorable for top-$K$ selection. The mask is applied to the scaled scores, excluding positions mapped to the stream minimum from the softmax normalization.
\end{enumerate}

\subsection{Query-Modulated Gating as a Zero-Shot Mixture-of-Experts}
\label{sec:gating}

To effectively route the query to the most appropriate information streams, we conceptualize our framework as a \textbf{Zero-Shot Mixture-of-Experts (MoE)} system. In traditional MoE architectures, a trainable gating network assigns weights to specialized expert modules. In Q-Gate, we leverage the zero-shot, in-context reasoning capabilities of a powerful LLM to act as an intelligent \textit{Gating Network}. Our three scoring streams ($S_g, S_m, S_c$) serve as the pretrained \textit{Modality Experts}. Instead of employing a static fusion rule, which is prone to the ``modal noise'' discussed earlier, the LLM Gater dynamically maps a query $q$ to a weight vector $W(q) = [w_g(q), w_m(q), w_c(q)]$, where $\sum w_i = 1$. This process is guided by a carefully designed prompt that encapsulates empirically-derived routing rules. The final score distribution is computed as the MoE's aggregated output:
\begin{equation}
    S_{final}(t) = \sum_{i \in \{g, m, c\}} w_i(q) \cdot S_i(t).
    \label{eq:moe_fusion}
\end{equation}
This zero-shot MoE formulation keeps Q-Gate optimization-free while retaining the dynamic, input-dependent adaptability of advanced routing networks.

\subsection{Sampling and Inference}
\label{ssec:sampling}
With the final score distribution $S_{final}$, we select the top-$K$ most relevant frames. A critical step for successful downstream reasoning is establishing a clear temporal bridge between the selected frames and their context. To solve the image-text misalignment issue, we explicitly format the input for the final VLM. Each selected frame is accompanied by its precise timestamp, formatted as \texttt{[Image at MM:SS]}. Similarly, any associated subtitles are also labeled with their timestamps. This structure enables the VLM to perform cross-verification by aligning evidence from different modalities along the common axis of time, improving its ability to answer complex, temporally-anchored questions.

\section{Experiments}
\label{sec:experiments}

We conduct extensive experiments to validate the effectiveness of our proposed \textbf{Q-Gate} framework. We aim to answer three key research questions: 
(1) Does our dynamic, multimodal keyframe selection strategy outperform representative single-modality and static-fusion baselines on long video question answering?
(2) How does Q-Gate perform across different video lengths and datasets?
(3) Is the performance gain attributable to our proposed query-aware gating mechanism?
We will address Q1 and Q2 in this section, while Q3 will be detailed in our ablation studies in Section~\ref{sec:ablation}.

\subsection{Experimental Setup}
\label{sec:setup}

\paragraph{Datasets.} We evaluate Q-Gate on two large-scale benchmarks that provide high-quality, synchronized multimodal data.

\textbf{LongVideoBench}~\cite{NEURIPS2024_329ad516} is a massive-scale benchmark with videos up to 60 minutes, categorized into three splits: \textit{Short} ($<$3\,min), \textit{Medium} (3--15\,min), and \textit{Long} (15--60\,min). \textbf{Video-MME}~\cite{Fu_2025_CVPR} is a comprehensive evaluation suite focused on multimodal reasoning. We primarily focus on their \textit{Long} and \textit{Medium} subsets to test long reasoning ability. We center our evaluation on these two benchmarks because their robust subtitle and audio-visual alignments are essential for probing the \textit{look-and-listen} synergy in human-centric narratives that motivates our design. We additionally evaluate Q-Gate on the subtitle-free MLVU benchmark~\cite{Zhou_2025_CVPR} (Section~\ref{ssec:mlvu}), where the Contextual Alignment stream should receive near-zero weight.

\paragraph{Baselines.} To situate Q-Gate within the current landscape, we compare it against four representative paradigms of keyframe selection, as detailed below:
\begin{itemize}
    \item \textbf{Uniform Sampling}: A content-agnostic reference that samples $K$ frames at fixed intervals.
    \item \textbf{AKS*}~\cite{Tang_2025_CVPR}: Represents the \textit{Global Matching} paradigm. We adopt a modified version (AKS*) by applying top-$K$ sampling on its relevance scores to ensure a strictly fixed frame budget $K$ for fair comparison, as the original adaptive partitioning can result in inconsistent frame counts.
    \item \textbf{VSLS}~\cite{NEURIPS2025_b432bc78}: Represents the \textit{Fine-grained Grounding} paradigm, utilizing object-level verification to refine frame importance.
    \item \textbf{T*}~\cite{Ye_2025_CVPR}: Represents the \textit{Iterative Search} paradigm, performing multi-step spatial-temporal zooming to localize task-relevant visual entities within the broader video context.
\end{itemize}
To isolate the contribution of our routing mechanism, all reproduced methods use the same downstream MLLM, frame budget, and evaluation protocol; each selector retains its method-specific scoring components.

\paragraph{Implementation Details.}
We set the keyframe budget to $K=8$ and $K=32$. Operating as an optimization-free routing module, Q-Gate leverages off-the-shelf pre-trained models as its expert streams: YOLO-World~\cite{Cheng_2024_CVPR} for Visual Grounding, BLIP~\cite{pmlr-v162-li22n} for Global Matching, and Sentence-BERT~\cite{reimers-gurevych-2019-sentence} for Contextual Alignment. The Query-Aware Gater is powered by GPT-4o~\cite{hurst2024gpt}, and we set the masked-softmax temperature to $\tau=0.5$. For downstream QA, we employ GPT-4o~\cite{hurst2024gpt} and Qwen3-VL-32B-Instruct~\cite{bai2025qwen3vltechnicalreport}.

\subsection{Main Results}
\label{ssec:main_results}

Table~\ref{tab:main_results} shows that Q-Gate outperforms all reproduced baselines in the majority of settings, with the largest gains on the challenging \textit{Long} and \textit{Medium} splits (e.g., a 6.40-point gain over AKS* on Video-MME \textit{Long} with Qwen3-VL at $K=32$), as its dynamic routing activates narrative context to resolve causal ambiguities that bottleneck purely visual methods. Using only $\mathbf{K=32}$ frames, Q-Gate attains $\mathbf{59.40\%}$ on LongVideoBench (\textit{Long}) and $\mathbf{61.19\%}$ on Video-MME (\textit{Long}), competitive with several reported 72B-parameter and API-based systems, although their backbones and evaluation settings are not directly comparable.

\begin{table*}[t]
  \centering
  \caption{\textbf{Downstream task evaluation on two benchmarks.} Black scores (\%) are from our controlled replication; \textcolor{blue}{blue} marks Q-Gate's absolute gain over the best reproduced baseline. Reported SOTA numbers are cited in \textcolor{gray}{gray} for reference only, as their backbones, parameter sizes, and frame inputs differ.}
  \label{tab:main_results}
  \renewcommand{\arraystretch}{1.05}
  \resizebox{\textwidth}{!}{%
  \begin{tabular}{@{} l c ccc || l c ccc @{}}
    \toprule
    \multicolumn{5}{c||}{\textbf{\textsc{LongVideoBench}}} & \multicolumn{5}{c}{\textbf{\textsc{Video-MME}}} \\
    \midrule
    \multirow{2}{*}{\textbf{Model and Strategy}} & \multirow{2}{*}{\textbf{Frame}} & \multicolumn{3}{c||}{\textbf{Video Length}} & \multirow{2}{*}{\textbf{Model and Strategy}} & \multirow{2}{*}{\textbf{Frame}} & \multicolumn{3}{c}{\textbf{Video Length}} \\
    & & \textbf{Long} & \textbf{Medium} & \textbf{Short} & & & \textbf{Long} & \textbf{Medium} & \textbf{Short} \\
    & & {\scriptsize 15-60m} & {\scriptsize 3-15m} & {\scriptsize <3m} & & & {\scriptsize >30min} & {\scriptsize 4-30min} & {\scriptsize <2min} \\
    \midrule

    GPT-4o (Uniform) & 8 & 40.43 & 43.69 & 57.65 & GPT-4o (Uniform) & 8 & 50.12 & 52.15 & 63.40 \\
    GPT-4o + AKS*~\cite{Tang_2025_CVPR} & 8 & 49.65 & 53.88 & 59.49 & GPT-4o + AKS*~\cite{Tang_2025_CVPR} & 8 & 54.00 & 57.86 & 66.85 \\
    GPT-4o + VSLS~\cite{NEURIPS2025_b432bc78} & 8 & 45.21 & 49.76 & 56.25 & GPT-4o + VSLS~\cite{NEURIPS2025_b432bc78} & 8 & 50.67 & 55.07 & 63.82 \\
    GPT-4o + T*~\cite{Ye_2025_CVPR} & 8 & 42.73 & 47.82 & 64.71 & GPT-4o + T*~\cite{Ye_2025_CVPR} & 8 & 51.26 & 53.58 & 60.36 \\
    \textbf{GPT-4o + Q-Gate (ours)} & 8 & \textbf{50.71} {\tiny \textcolor{blue}{(+1.06)}} & \textbf{56.55} {\tiny \textcolor{blue}{(+2.67)}} & 65.41 {\tiny \textcolor{blue}{(+0.70)}} & \textbf{GPT-4o + Q-Gate (ours)} & 8 & \textbf{54.78} {\tiny \textcolor{blue}{(+0.78)}} & \textbf{59.68} {\tiny \textcolor{blue}{(+1.82)}} & \textbf{67.16} {\tiny \textcolor{blue}{(+0.31)}} \\
    \midrule
    
    GPT-4o (Uniform) & 32 & 43.79 & 46.36 & 60.00 & GPT-4o (Uniform) & 32 & 52.40 & 52.96 & 68.63 \\
    GPT-4o + AKS*~\cite{Tang_2025_CVPR} & 32 & 47.70 & 50.97 & 61.88 & GPT-4o + AKS*~\cite{Tang_2025_CVPR} & 32 & 52.89 & 59.20 & 66.63 \\
    GPT-4o + VSLS~\cite{NEURIPS2025_b432bc78} & 32 & 44.50 & 49.51 & \textbf{68.24} & GPT-4o + VSLS~\cite{NEURIPS2025_b432bc78} & 32 & 50.68 & 55.11 & 64.05 \\
    GPT-4o + T*~\cite{Ye_2025_CVPR} & 32 & 43.62 & 48.79 & 63.53 & GPT-4o + T*~\cite{Ye_2025_CVPR} & 32 & 51.48 & 56.01 & 66.28 \\
    \textbf{GPT-4o + Q-Gate (ours)} & 32 & \textbf{50.35} {\tiny \textcolor{blue}{(+2.65)}} & \textbf{53.64} {\tiny \textcolor{blue}{(+2.67)}} & 63.75 & \textbf{GPT-4o + Q-Gate (ours)} & 32 & \textbf{60.05} {\tiny \textcolor{blue}{(+7.16)}} & \textbf{62.77} {\tiny \textcolor{blue}{(+3.57)}} & \textbf{69.44} {\tiny \textcolor{blue}{(+0.81)}} \\
    \midrule\midrule
    
    Qwen3-VL (Uniform) & 8 & 46.63 & 51.46 & \textbf{71.76} & Qwen3-VL (Uniform) & 8 & 51.14 & 52.15 & 63.40 \\
    Qwen3-VL + AKS*~\cite{Tang_2025_CVPR} & 8 & 54.61 & 60.92 & 70.59 & Qwen3-VL + AKS*~\cite{Tang_2025_CVPR} & 8 & 54.45 & 58.33 & \textbf{68.79} \\
    Qwen3-VL + VSLS~\cite{NEURIPS2025_b432bc78} & 8 & 51.60 & 55.83 & 69.41 & Qwen3-VL + VSLS~\cite{NEURIPS2025_b432bc78} & 8 & 51.37 & 53.76 & 66.50 \\
    Qwen3-VL + T*~\cite{Ye_2025_CVPR} & 8 & 50.89 & 54.13 & 67.06 & Qwen3-VL + T*~\cite{Ye_2025_CVPR} & 8 & 52.28 & 55.60 & 65.79 \\
    \textbf{Qwen3-VL + Q-Gate (ours)} & 8 & \textbf{58.69} {\tiny \textcolor{blue}{(+4.08)}} & \textbf{63.11} {\tiny \textcolor{blue}{(+2.19)}} & 68.24 & \textbf{Qwen3-VL + Q-Gate (ours)} & 8 & \textbf{57.19} {\tiny \textcolor{blue}{(+2.74)}} & \textbf{61.29} {\tiny \textcolor{blue}{(+2.96)}} & \textbf{73.04} {\tiny \textcolor{blue}{(+4.25)}}  \\
    \midrule
    
    Qwen3-VL (Uniform) & 32 & 51.24 & 56.55 & \textbf{71.76} & Qwen3-VL (Uniform) & 32 & 51.60 & 54.57 & 65.85 \\
    Qwen3-VL + AKS*~\cite{Tang_2025_CVPR} & 32 & 57.80 & \textbf{65.29} & 67.06 & Qwen3-VL + AKS*~\cite{Tang_2025_CVPR} & 32 & 54.79 & 62.90 & 75.68 \\
    Qwen3-VL + VSLS~\cite{NEURIPS2025_b432bc78} & 32 & 52.66 & 58.74 & 65.88 & Qwen3-VL + VSLS~\cite{NEURIPS2025_b432bc78} & 32 & 54.68 & 59.81 & 74.35 \\
    Qwen3-VL + T*~\cite{Ye_2025_CVPR} & 32 & 51.95 & 56.55 & 69.41 & Qwen3-VL + T*~\cite{Ye_2025_CVPR} & 32 & 54.22 & 58.57 & 75.00 \\
    \textbf{Qwen3-VL + Q-Gate (ours)} & 32 & \textbf{59.40} {\tiny \textcolor{blue}{(+1.60)}} & 63.11 & 70.59 & \textbf{Qwen3-VL + Q-Gate (ours)} & 32 & \textbf{61.19} {\tiny \textcolor{blue}{(+6.40)}} & \textbf{66.13} {\tiny \textcolor{blue}{(+3.23)}} & \textbf{79.41} {\tiny \textcolor{blue}{(+3.73)}} \\
    \midrule\midrule
    
    \textcolor{gray}{FOCUS (LLaVA-V-7B)~\cite{zhu2026focus}} & \textcolor{gray}{64} & \textcolor{gray}{63.7} & \textcolor{gray}{59.0} & \textcolor{gray}{72.3} & \textcolor{gray}{FOCUS (LLaVA-V-7B)~\cite{zhu2026focus}} & \textcolor{gray}{64} & \textcolor{gray}{56.1} & \textcolor{gray}{63.5} & \textcolor{gray}{76.5} \\
    
    \textcolor{gray}{LLaVA-OneVision-72B~\cite{li2025llavaonevision}} & \textcolor{gray}{32} & \textcolor{gray}{59.3} & \textcolor{gray}{63.9} & \textcolor{gray}{77.4} & \textcolor{gray}{LLaVA-OneVision-72B~\cite{li2025llavaonevision}} & \textcolor{gray}{32} & \textcolor{gray}{60.0} & \textcolor{gray}{62.2} & \textcolor{gray}{66.3} \\
    
    \textcolor{gray}{GPT-4o (0513)~\cite{hurst2024gpt}} & \textcolor{gray}{256} & \textcolor{gray}{61.6} & \textcolor{gray}{66.7} & \textcolor{gray}{76.8} & \textcolor{gray}{Q-Frame (GPT-4o)~\cite{Zhang_2025_ICCV}} & \textcolor{gray}{8} & \textcolor{gray}{57.6} & \textcolor{gray}{63.8} & \textcolor{gray}{69.9} \\
    
    \textcolor{gray}{LLaVA-Video-72B-Qwen2~\cite{zhang2025llavavideo}} & \textcolor{gray}{128} & \textcolor{gray}{59.3} & \textcolor{gray}{63.9} & \textcolor{gray}{77.4} & \textcolor{gray}{Gemini-1.5-Pro (0615)~\cite{Fu_2025_CVPR}} & \textcolor{gray}{1 fps} & \textcolor{gray}{67.4} & \textcolor{gray}{74.3} & \textcolor{gray}{75.0} \\
    \bottomrule
  \end{tabular}%
  }
\end{table*}

\section{Ablation Studies and Analysis}
\label{sec:ablation}

\subsection{Impact of Multi-Granularity Modalities}
\label{ssec:ablation_modalities}

To evaluate the necessity of each scoring stream, we ablate them
individually using Qwen3-VL-32B-Instruct ($K=32$). As shown in
Table~\ref{tab:ablation_modalities}, the Full Model achieves the best overall balance, confirming that the three granularities are
complementary. Removing \textbf{Contextual Alignment ($S_c$)} causes a substantial drop from 59.40\% to 54.08\% on the \textit{Long} split of LongVideoBench, validating that visual signals alone are insufficient for narrative-heavy content and causality. The contributions of $S_g$ and $S_m$ are task-dependent: $S_g$ is more critical for detail-sensitive queries, while $S_m$ dominates for broader scenes. Omitting $S_m$ on LongVideoBench \textit{Medium} even yields a marginal gain, suggesting global matching can introduce cross-scene ambiguity and further motivating dynamic over static fusion. Synergizing all three streams gives the most robust performance across video lengths and question types.

\begin{table}[t]
  \centering
  \caption{Ablation of the three scoring streams with Qwen3-VL-32B-Instruct ($K=32$). The full model provides the best overall balance, while removing Contextual Alignment causes the largest drop on LongVideoBench Long.}
  \label{tab:ablation_modalities}
  \renewcommand{\arraystretch}{1.0}
  \resizebox{\columnwidth}{!}{%
  \begin{tabular}{l|ccc|ccc}
    \toprule
    \multirow{2}{*}{\textbf{Model Variant}} & \multicolumn{3}{c|}{\textbf{LongVideoBench} (\%)} & \multicolumn{3}{c}{\textbf{Video-MME} (\%)} \\
    \cmidrule(lr){2-4} \cmidrule(lr){5-7}
    & \textbf{Long} & \textbf{Med.} & \textbf{Short} & \textbf{Long} & \textbf{Med.} & \textbf{Short} \\
    \midrule
    \textbf{Full Model (Q-Gate)} & \textbf{59.40} & 63.11 & \textbf{70.59} & \textbf{61.19} & 66.13 & \textbf{79.41} \\
    \midrule
    w/o Contextual Align. ($S_c$) & 54.08 & 58.01 & 67.06 & 58.79 & 63.17 & 76.31 \\
    w/o Visual Grounding ($S_g$) & 58.33 & 61.89 & 64.71 & 59.36 & \textbf{66.40} & 78.43 \\
    w/o Global Matching ($S_m$) & 57.45 & \textbf{64.81} & 63.53 & 60.73 & 65.32 & 78.59 \\
    \bottomrule
  \end{tabular}%
  }
\end{table}

\subsection{Efficacy of Query-Aware Gating}
\label{ssec:ablation_gating}

To verify that dynamic gating suppresses modality-specific noise, we compare Q-Gate against a \textbf{static} fusion baseline (Equal Weights: $w_g=w_m=w_c=1/3$). We further include a rule-based \textbf{heuristic} gater (handcrafted keyword rules) as a diagnostic probe on LongVideoBench to test whether intent-to-modality routing can be reduced to surface pattern matching. As shown in Table~\ref{tab:ablation_gating}, the heuristic gater fails to surpass even static fusion (44.40\% vs.\ 47.70\% on \textit{Long}), indicating that this routing requires semantic reasoning rather than keyword matching. Our dynamic routing instead yields gains in most settings, most pronounced on Qwen3-VL-32B-Instruct at $K=32$, where the static baseline drops to 55.67\% on \textit{Short} videos while Q-Gate reaches 70.59\%, yielding a 14.92-point gain. This supports our \textbf{``modal noise''} insight: fusing irrelevant subtitles into visual-centric queries induces cross-modal negative transfer, which Q-Gate mitigates by assigning near-zero weight to the narrative stream. Static fusion stays competitive only on the balanced \textit{Medium} splits, while Q-Gate dominates the \textit{Long} and \textit{Short} splits.

\begin{table}[t]
  \centering
  \caption{Ablation study of gating strategies across different VLMs and frame budgets ($K$). We compare Static fusion and a rule-based Heuristic gater against our Dynamic Query-Aware Gating. The Heuristic gater is evaluated only as a diagnostic probe on LongVideoBench with GPT-4o at $K=8$; dashes indicate settings outside the scope of this probe.}
  \label{tab:ablation_gating}
  \renewcommand{\arraystretch}{1.0}
  \resizebox{\columnwidth}{!}{%
  \begin{tabular}{lcc|ccc|ccc}
    \toprule
    \multirow{2}{*}{\textbf{VLM}} & \multirow{2}{*}{\textbf{K}} & \multirow{2}{*}{\textbf{Strategy}} & \multicolumn{3}{c|}{\textbf{LongVideoBench} (\%)} & \multicolumn{3}{c}{\textbf{Video-MME} (\%)} \\
    \cmidrule(lr){4-6} \cmidrule(lr){7-9}
    & & & \textbf{Long} & \textbf{Med.} & \textbf{Short} & \textbf{Long} & \textbf{Med.} & \textbf{Short} \\
    \midrule
    \multirow{3}{*}{GPT-4o} & \multirow{3}{*}{8} & Static & 47.70 & 53.64 & 55.56 & 54.44 & \textbf{60.00} & 66.67 \\
     & & Heuristic & 44.40 & 52.43 & 56.33 & -- & -- & -- \\
     & & \textbf{Dynamic} & \textbf{50.71} & \textbf{56.55} & \textbf{65.41} & \textbf{54.78} & 59.68 & \textbf{67.16} \\
    \midrule
    \multirow{2}{*}{Qwen3-VL} & \multirow{2}{*}{32} & Static & 57.73 & 62.62 & 55.67 & 60.05 & \textbf{66.80} & 78.92 \\
     & & \textbf{Dynamic} & \textbf{59.40} & \textbf{63.11} & \textbf{70.59} & \textbf{61.19} & 66.13 & \textbf{79.41} \\
    \bottomrule
  \end{tabular}%
  }
\end{table}

\subsection{Robustness to the Gater Model}
\label{ssec:ablation_gater}

To assess Q-Gate's robustness across Gater models, we test whether a locally deployable open-source model can replace the proprietary LLM as the ``strategy gater.'' We substitute GPT-4o with \textbf{Qwen3-VL-32B-Instruct} for weight estimation while keeping the downstream QA backbone fixed. As shown in Table~\ref{tab:ablation_gater}, GPT-4o holds a distinct advantage at $K=8$, but the gap narrows sharply at $K=32$, where the Qwen3-VL-gated variant is highly competitive. Q-Gate's routing is thus not tethered to proprietary models, letting users trade peak performance for cost-efficient, privacy-oriented local deployment without substantial degradation.

We further examine two potential confounders. First, to confirm that the gain does not come from the gater observing the answer options, we re-run the gating with options removed, so that Q-Gate accesses the same selection-time information as query-only baselines such as AKS* and T*. On a cross-family downstream model (InternVL3-8B~\cite{zhu2025internvl3exploringadvancedtraining}, mean of five runs), this query-only variant still surpasses the AKS* baseline by 0.63 points overall and 1.89 points on \textit{Long} videos, while access to the answer options contributes only 0.18 points, indicating that the improvement originates from multimodal routing rather than from the answer choices. Second, to check that the generated routing weights are meaningful, we compare the gater's top-1 modality against a category-derived control set on cleanly mappable questions, obtaining an agreement of 87.1\% for the contextual stream and 98.1\% for the matching stream. On 200-sample LongVideoBench subsets ($K=8$), paraphrasing or abstracting the queries reduces overall accuracy by 2.17 points, while dropping 30\% of subtitle words causes a 3.50-point decline. Q-Gate nevertheless remains 14.0 points above a subtitle-only top-$K$ selector, showing that its visual streams provide non-redundant evidence and that routing remains stable under moderate query and subtitle noise, though it still inherits errors from its expert streams.

\begin{table}[t]
  \centering
  \caption{Performance comparison using different models as the Gater. Experiments are on LongVideoBench and Video-MME with Qwen3-VL-32B-Instruct as the downstream QA model.}
  \label{tab:ablation_gater}
  \renewcommand{\arraystretch}{1.0}
  \resizebox{\columnwidth}{!}{%
  \begin{tabular}{lc|ccc|ccc}
    \toprule
    \multirow{2}{*}{\textbf{K}} & \multirow{2}{*}{\textbf{Gater Model}} & \multicolumn{3}{c|}{\textbf{LongVideoBench} (\%)} & \multicolumn{3}{c}{\textbf{Video-MME} (\%)} \\
    \cmidrule(lr){3-5} \cmidrule(lr){6-8}
    & & \textbf{Long} & \textbf{Med.} & \textbf{Short} & \textbf{Long} & \textbf{Med.} & \textbf{Short} \\
    \midrule
    \multirow{2}{*}{8} & GPT-4o & \textbf{58.69} & \textbf{63.11} & \textbf{68.24} & \textbf{57.19} & \textbf{61.29} & \textbf{73.04} \\
                       & Qwen3-VL & 56.38 & 61.17 & 63.53 & 53.77 & 61.02 & 72.39 \\
    \midrule
    \multirow{2}{*}{32} & GPT-4o & \textbf{59.40} & 63.11 & \textbf{70.59} & \textbf{61.19} & \textbf{66.13} & \textbf{79.41} \\
                        & Qwen3-VL & 58.16 & \textbf{63.35} & 65.88 & 59.47 & 64.78 & 78.95 \\
    \bottomrule
  \end{tabular}%
  }
\end{table}

\subsection{Generalization to a Subtitle-Free Benchmark}
\label{ssec:mlvu}

The design of Q-Gate targets the look-and-listen synergy, which raises a natural question: does it degrade when narrative cues are unavailable? We evaluate this on MLVU~\cite{Zhou_2025_CVPR}, a purely visual long-video benchmark without subtitles, using GPT-4o as the downstream model. As reported in Table~\ref{tab:mlvu}, Q-Gate still outperforms both baselines, VSLS and AKS*, by 2.86 points overall at $K=8$ and 1.35 points overall at $K=32$. In the absence of subtitles, the Contextual Alignment stream automatically receives near-zero weight and Q-Gate reduces to a strong visual selector, degrading gracefully rather than breaking, consistent with the ``suppress modal noise'' principle: the gater allocates weight only to the streams that carry usable signal for a given query.

\begin{table}[t]
  \centering
  \caption{Results on MLVU~\cite{Zhou_2025_CVPR}, a subtitle-free benchmark (GPT-4o downstream). Even without narrative cues, Q-Gate exceeds both single-stream baselines, confirming graceful degradation to a strong visual selector.}
  \label{tab:mlvu}
  \renewcommand{\arraystretch}{1.0}
  \resizebox{\columnwidth}{!}{%
  \begin{tabular}{l|ccc|ccc}
    \toprule
    \multirow{2}{*}{\textbf{Strategy}} & \multicolumn{3}{c|}{\textbf{MLVU} ($K=8$)} & \multicolumn{3}{c}{\textbf{MLVU} ($K=32$)} \\
    \cmidrule(lr){2-4} \cmidrule(lr){5-7}
    & \textbf{Med.} & \textbf{Long} & \textbf{Ov.} & \textbf{Med.} & \textbf{Long} & \textbf{Ov.} \\
    \midrule
    VSLS~\cite{NEURIPS2025_b432bc78} & 57.86 & 60.00 & 58.29 & 59.71 & 68.57 & 61.49 \\
    AKS*~\cite{Tang_2025_CVPR} & 56.43 & \textbf{71.43} & 59.43 & 60.43 & 77.14 & 63.79 \\
    \textbf{Q-Gate (ours)} & \textbf{60.00} & \textbf{71.43} & \textbf{62.29} & \textbf{61.43} & \textbf{80.00} & \textbf{65.14} \\
    \bottomrule
  \end{tabular}%
  }
\end{table}

\begin{figure*}[tbp]
  \centering
  \includegraphics[width=\textwidth]{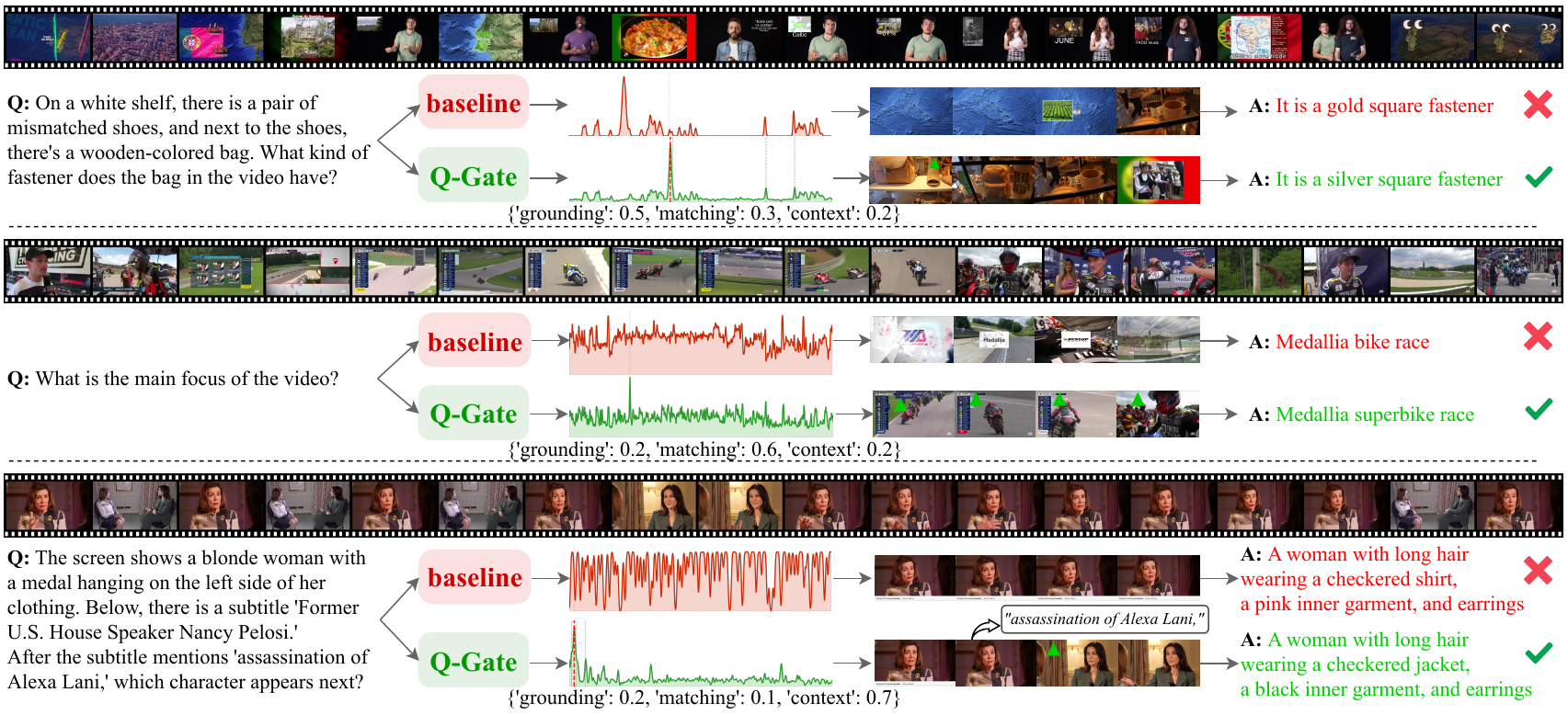}
  \caption{Qualitative visualization of Q-Gate's query-adaptive routing. \textbf{Top:} a detail query weights grounding to pinpoint the ``fastener.'' \textbf{Middle:} a thematic query weights matching for the ``superbike'' context. \textbf{Bottom:} a subtitle-anchored query shifts to context via dialogue cues to recover the correct character.
  \framecredit{
   LongVideoBench~\cite{NEURIPS2024_329ad516},
   video IDs \texttt{mq6L8CnNJXc} and \texttt{WaiGdRYD36k};
   Video-MME~\cite{Fu_2025_CVPR},
   video ID \texttt{1wzgMHrkrys}
  }
  }
  \Description{Three example rows, each showing a query, the per-frame scores of the three streams for the baseline versus Q-Gate, the selected keyframes, and the resulting answer, illustrating how Q-Gate reweights streams by query type.}
  \label{fig:qualitative_vis}
\end{figure*}

\subsection{Interpretability of Query-Aware Gating}
\label{ssec:interpretability}

\begin{figure}[tbp]
  \centering

  \includegraphics[width=0.49\linewidth]
  {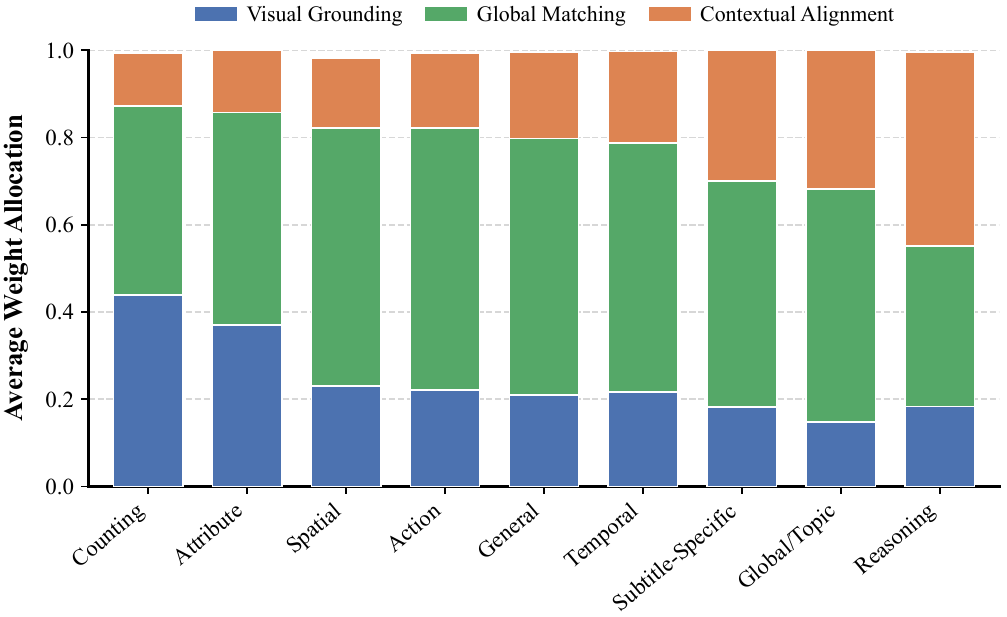}
  \hfill
  \includegraphics[width=0.49\linewidth]
  {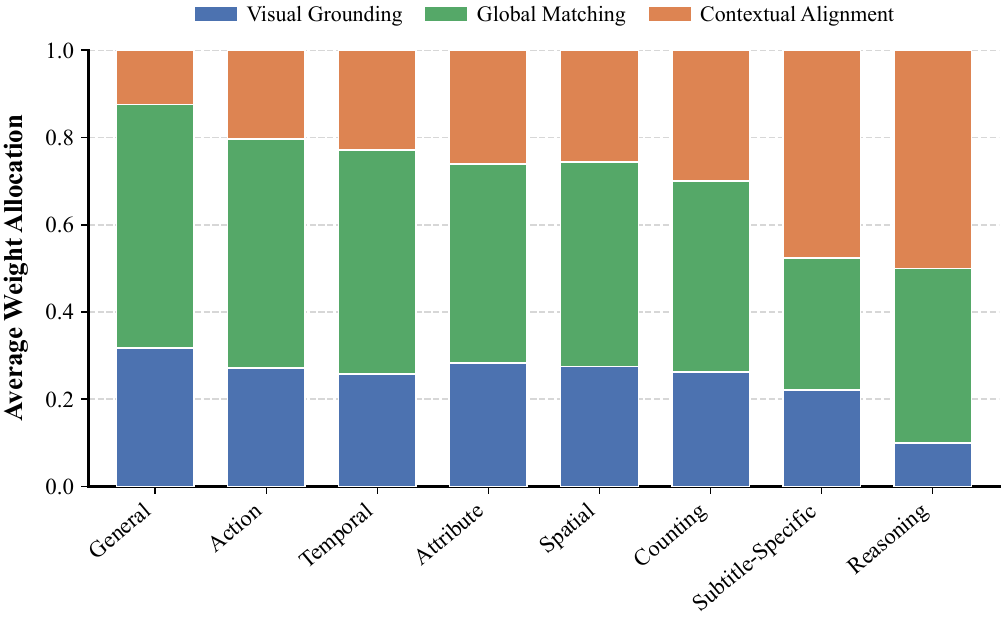}

  {\footnotesize\textbf{(a)} Query-category weight allocation}

  \includegraphics[width=0.98\linewidth]
  {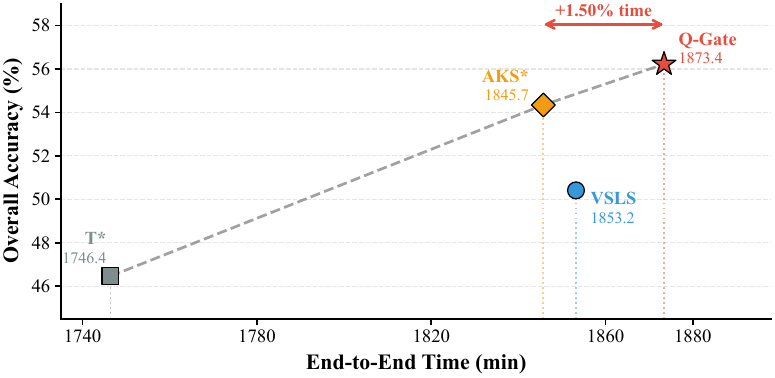}

  {\footnotesize\textbf{(b)} Accuracy--latency trade-off}

  \caption{
  Interpretability and efficiency analyses.
  \textbf{(a)} Q-Gate shifts from visual-heavy to contextual
  streams as query abstraction increases.
  \textbf{(b)} Q-Gate improves accuracy over AKS* with only
  1.50\% additional end-to-end time.
  }

  \Description{
  Two panels summarize Q-Gate's behavior. Panel (a) contains
  stacked bar charts of the average grounding, matching, and
  contextual weights across question categories on LongVideoBench
  and Video-MME. Panel (b) compares T*, VSLS, AKS*, and Q-Gate
  by overall accuracy and end-to-end processing time.
  }

  \label{fig:interpretability_efficiency}
\end{figure}

To assess whether Q-Gate routes queries in a principled way, we analyze its average weight allocation across question categories on both benchmarks (Figure~\ref{fig:interpretability_efficiency}(a)). The distributions match human intuition: detail-oriented tasks (e.g., \textit{Counting}, \textit{Attribute}) assign prominent weight to \textbf{Visual Grounding}, broader queries (e.g., \textit{Action}) favor \textbf{Global Matching}, and abstract, plot-driven queries (e.g., \textit{Reasoning}, \textit{Subtitle-Specific}) raise \textbf{Contextual Alignment} to nearly half of the allocation, making Q-Gate's routing decisions directly interpretable.

\subsection{Qualitative Analysis}
\label{ssec:qualitative}

Figure~\ref{fig:qualitative_vis} illustrates Q-Gate's routing across three scenarios. A fine-grained ``fastener'' query prioritizes grounding ($w_g{=}0.5$), sharpening the scattered baseline noise into a peak at the target frame; a thematic ``main focus'' query shifts to matching ($w_m{=}0.6$) to isolate the ``superbike'' context; and a subtitle-anchored query, where the static baseline fails as a ``deaf observer,'' ``listens'' via context ($w_c{=}0.7$), spiking right after the textual cue to locate the correct character.

\begin{figure}[tbp]
  \centering
  \includegraphics[width=\columnwidth]
  {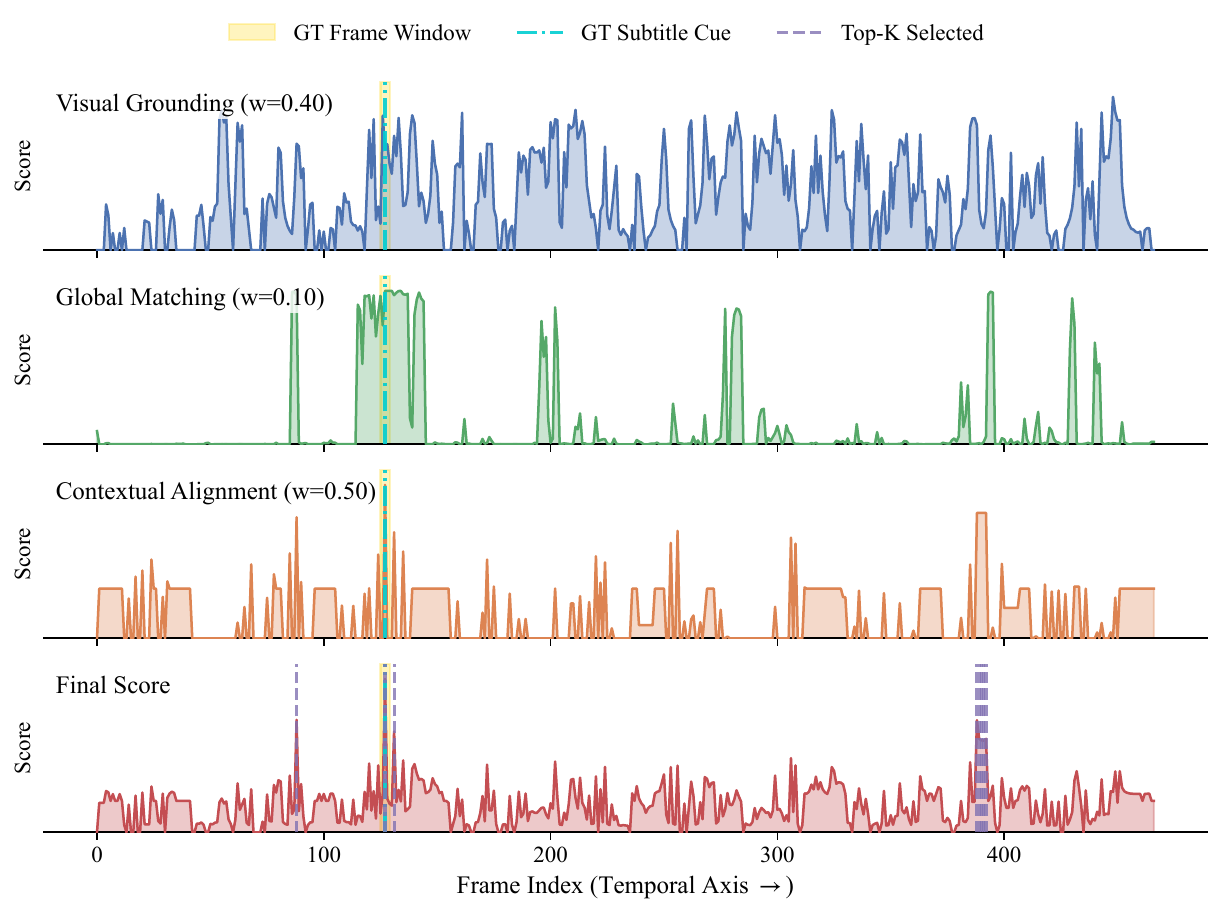}
  \caption{
  Internal fusion for a subtitle-anchored query.
  Contextual Alignment localizes the relevant interval, while
  Visual Grounding refines the evidence; the fused top-$K$ frames
  align with the ground-truth window.
  }
  \Description{
  Four temporally aligned score plots show Visual Grounding,
  Global Matching, Contextual Alignment, and the final fused score.
  The contextual score identifies the relevant interval, and the
  fused top-K selections overlap the annotated ground-truth window.
  }
  \label{fig:fusion_dynamics}
\end{figure}

\subsection{Internal Fusion Dynamics and Alignment}
\label{ssec:internal_dynamics}

Figure~\ref{fig:fusion_dynamics} visualizes Q-Gate's score fusion for the cross-modal query in Figure~\ref{fig:framework}, which requires temporal localization (\textit{``When the subtitle says...''}) and fine-grained perception (\textit{``what color...''}). Q-Gate assigns most weight to Contextual Alignment ($w_c=0.50$) and Visual Grounding ($w_g=0.40$). Although recurring elements make the visual stream noisy, contextual evidence anchors the relevant interval, and the fused top-$K$ frames align with the ground-truth window, confirming correct temporal localization.

\subsection{Efficiency and End-to-End Latency Analysis}
\label{ssec:latency_analysis}

We measure end-to-end latency on LongVideoBench ($K=8$).
As shown in Figure~\ref{fig:interpretability_efficiency}(b),
Q-Gate improves accuracy over AKS* with only 1.50\% additional
processing time. The three expert streams are evaluated once per video--question pair and run independently in parallel, so their wall-clock latency is set by the slowest stream rather than the sum of all three, and the text-only Gater adds only a small routing cost.

\section{Conclusion}
\label{sec:conclusion}

We proposed \textbf{Q-Gate}, a plug-and-play, training-free framework that treats keyframe selection as query-conditioned modality routing. Through its Query-Modulated Gating Mechanism, Q-Gate adaptively decides when to prioritize visual evidence and when to rely on narrative context, suppressing irrelevant modal signals. Extensive experiments demonstrate strong gains in accuracy, interpretability, and efficiency across multiple benchmarks, backbones, and frame budgets. These results suggest that adaptive routing is a promising direction for robust and scalable long video understanding.

\clearpage

\begin{acks}
The authors acknowledge the high-performance computing resources provided by The Hong Kong University of Science and Technology (Guangzhou), the GPU resources provided by Prof. Hui Xiong's research group, and access to large language model APIs provided by the College of Future Technology at The Hong Kong University of Science and Technology (Guangzhou).
\end{acks}

\bibliographystyle{ACM-Reference-Format}
\balance
\bibliography{sample-base}

\end{document}